\documentclass{article}

\usepackage[utf8]{inputenc} 
\usepackage[T1]{fontenc}    
\usepackage{hyperref}       
\usepackage{url}            
\usepackage{booktabs}       
\usepackage{amsfonts}       
\usepackage{nicefrac}       
\usepackage{microtype}      
\usepackage{authblk}

\usepackage{xcolor}
\usepackage{url}
\usepackage{latexsym}

\usepackage{mathtools}
\usepackage{calc}
\usepackage{amsmath, amsthm, amssymb, amsfonts}
\usepackage{naaclhlt2018}

\usepackage{graphicx}
\def\shrug{\texttt{\raisebox{0.75em}{\char`\_}\char`\\\char`\_\kern-0.5ex(\kern-0.25ex\raisebox{0.25ex}{\rotatebox{45}{\raisebox{-.75ex}"\kern-1.5ex\rotatebox{-90})}}\kern-0.5ex)\kern-0.5ex\char`\_/\raisebox{0.75em}{\char`\_}}}

\aclfinalcopy

\title{Simultaneously Self-Attending to All Mentions for \\Full-Abstract Biological Relation Extraction}

\author{\textbf{Patrick Verga,  Emma Strubell,  Andrew McCallum} \\
  College of Information and Computer Sciences \\
  University of Massachusetts Amherst \\
  {\tt \{pat, strubell, mccallum\}@cs.umass.edu}} 

\date{}

\begin{document}
\maketitle
\begin{abstract}
Most work in relation extraction forms a prediction by looking at a short span of text within a single sentence containing a single entity pair mention. This approach often does not consider interactions across mentions, requires redundant computation for each mention pair, and ignores relationships expressed across sentence boundaries. These problems are exacerbated by the document- (rather than sentence-) level annotation common in biological text. In response, we propose a model which simultaneously predicts relationships between all mention pairs in a document. We form pairwise predictions over entire paper abstracts using an efficient self-attention encoder. All-pairs mention scores allow us to perform multi-instance learning by aggregating over mentions to form entity pair representations. We further adapt to settings without mention-level annotation by jointly training to predict named entities and adding a corpus of weakly labeled data. In experiments on two Biocreative benchmark datasets, we achieve state of the art performance on the Biocreative V Chemical Disease Relation dataset for models without external KB resources. We also introduce a new dataset an order of magnitude larger than existing human-annotated biological information extraction datasets and more accurate than distantly supervised alternatives.

\end{abstract}

\section{Introduction}
\label{sec:intro}
With few exceptions \citep{swampillai-stevenson:2011:RANLP,quirk-poon:2017:EACLlong, peng2017cross}, nearly all work in relation extraction focuses on classifying a short span of text within a single sentence containing a single entity pair mention. However, relationships between entities are often expressed across sentence boundaries or otherwise require a larger context to disambiguate. For example, 30\% of relations in the Biocreative V CDR dataset (\S \ref{sec:cdr_data}) are expressed across sentence boundaries, such as in the following excerpt expressing a relationship between the chemical \textbf{\textcolor{blue}{azathioprine}} and the disease \textbf{\textcolor{red}{fibrosis}}:

\begin{quote}
\small
\textit{Treatment of psoriasis with \textbf{\textcolor{blue}{azathioprine}}}. \textbf{\textcolor{blue}{Azathioprine}} treatment benefited 19 (66\%) out of 29 patients suffering from severe psoriasis. Haematological complications were not troublesome and results of biochemical liver function tests remained normal. Minimal cholestasis was seen in two cases and portal \textbf{\textcolor{red}{fibrosis}} of a reversible degree in eight. Liver biopsies should be undertaken at regular intervals if \textbf{\textcolor{blue}{azathioprine}} therapy is continued so that structural liver damage may be detected at an early and reversible stage.
\end{quote}
Though the entities' mentions never occur in the same sentence, the above example expresses that the chemical entity \emph{azathioprine} can cause the side effect \emph{fibrosis}. Relation extraction models which consider only within-sentence relation pairs cannot extract this fact without knowledge of the complicated coreference relationship between \emph{eight} and \emph{azathioprine treatment}, which, without features from a complicated pre-processing pipeline, cannot be learned by a model which considers entity pairs in isolation. Making separate predictions for each mention pair also obstructs multi-instance learning \citep{riedel2010modeling, surdeanu2012multi}, a technique which aggregates entity representations from mentions in order to improve robustness to noise in the data. Like the majority of relation extraction data, most annotation for biological relations is distantly supervised, and so we could benefit from a model which is amenable to multi-instance learning.  


In addition to this loss of cross-sentence and cross-mention reasoning capability, traditional mention pair relation extraction models typically introduce computational inefficiencies by independently extracting features for and scoring every pair of mentions, even when those mentions occur in the same sentence and thus could share representations. In the CDR training set, this requires separately encoding and classifying each of the 5,318 candidate mention pairs independently, versus encoding each of the 500 abstracts once. Though abstracts are longer than e.g. the text between mentions, many sentences contain multiple mentions, leading to redundant computation.

However, encoding long sequences in a way which effectively incorporates long-distance context can be prohibitively expensive. Long Short Term Memory (LSTM) networks \citep{hochreiter1997long} are among the most popular token encoders due to their capacity to learn high-quality representations of text, but their ability to leverage the fastest computing hardware is thwarted due to their computational dependence on the length of the sequence --- each token's representation requires as input the representation of the previous token, limiting the extent to which computation can be parallelized. Convolutional neural networks (CNNs), in contrast, can be executed entirely in parallel across the sequence, but the amount of context incorporated into a single token's representation is limited by the depth of the network, and very deep networks can be difficult to learn \citep{hochreiter1998vanishing}. These problems are exacerbated by longer sequences, limiting the extent to which previous work explored full-abstract relation extraction. 

To facilitate efficient full-abstract relation extraction from biological text, we propose Bi-affine Relation Attention Networks (BRANs), a combination of network architecture, multi-instance and multi-task learning designed to extract relations between entities in biological text without requiring explicit mention-level annotation. We synthesize convolutions and self-attention, a modification of the Transformer encoder introduced by \citet{vaswani2017attention}, over sub-word tokens to efficiently incorporate into token representations rich context between distant mention pairs across the entire abstract. We score all pairs of mentions in parallel using a bi-affine operator, and aggregate over mention pairs using a soft approximation of the max function in order to perform multi-instance learning. We jointly train the model to predict relations and entities, further improving robustness to noise and lack of gold annotation at the mention level. 

In extensive experiments on two benchmark biological relation extraction datasets, we achieve state of the art performance for a model using no external knowledge base resources in experiments on the Biocreative V CDR dataset, and outperform comparable baselines on the Biocreative VI ChemProt dataset. We also introduce a new dataset which is an order of magnitude larger than existing gold-annotated biological relation extraction datasets while covering a wider range of entity and relation types and with higher accuracy than distantly supervised datasets of the same size. We provide a strong baseline on this new dataset, and encourage its use as a benchmark for future biological relation extraction systems.\footnote{Our code and data are publicly available at: \protect\url{https://github.com/patverga/bran}.}

\section{Model}
\label{sec:model}
We designed our model to efficiently encode long contexts spanning multiple sentences while forming pairwise predictions without the need for mention pair-specific features. To do this, our model first encodes input token embeddings using self-attention. These embeddings are used to predict both entities and relations. The relation extraction module converts each token to a \emph{head} and \emph{tail} representation. These representations are used to form mention pair predictions using a bi-affine operation with respect to learned relation embeddings. Finally, these mention pair predictions are pooled to form entity pair predictions, expressing whether each relation type is expressed by each relation pair.

\begin{figure}[t!]
  \begin{minipage}[c]{0.45\textwidth}
    \includegraphics[scale=.8]{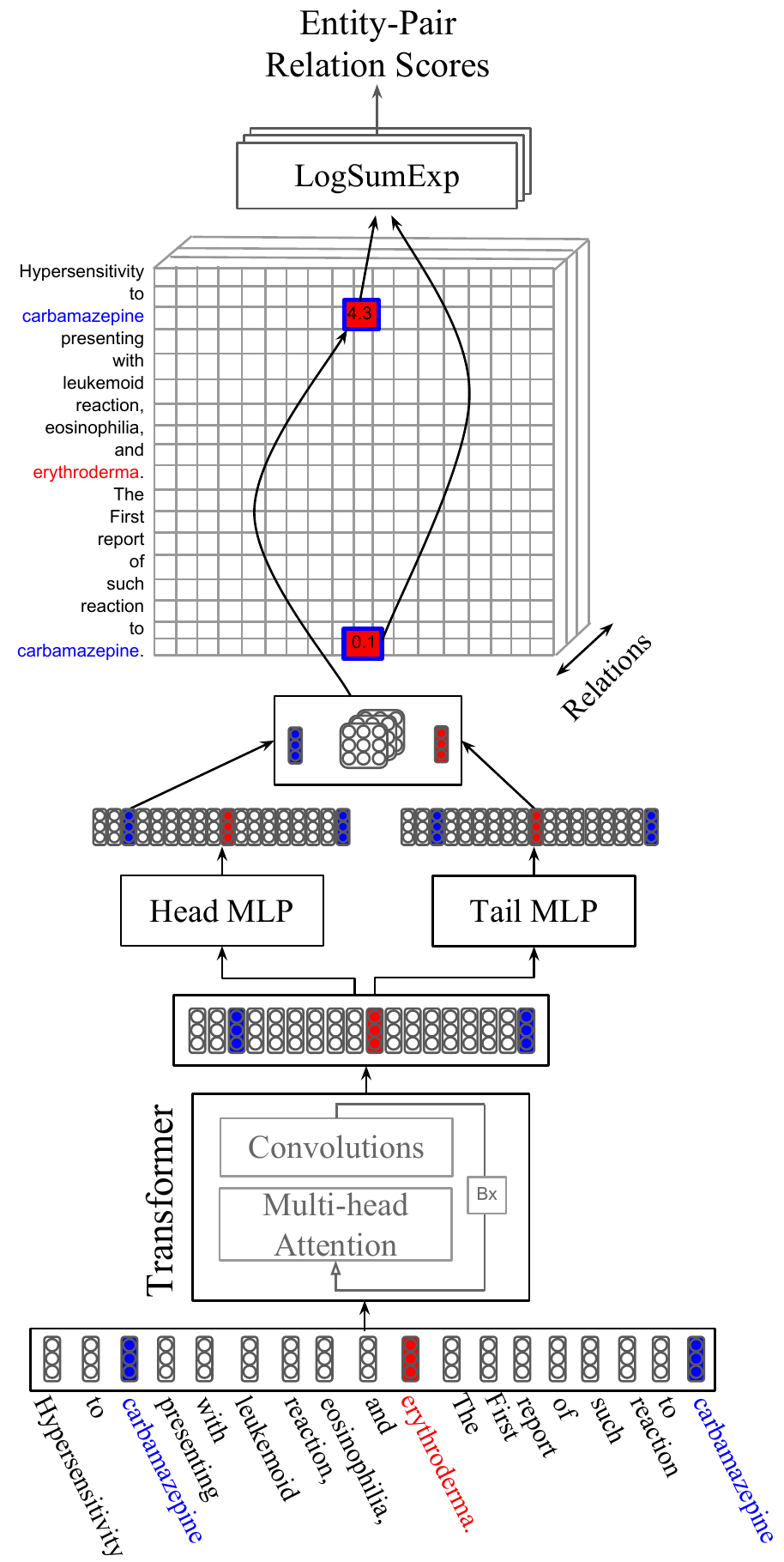}
  \end{minipage}\hfill
  \begin{minipage}[c]{0.45\textwidth}
    \caption{The relation extraction architecture. Inputs are contextually encoded using the Transformer\citep{vaswani2017attention}, made up of $B$ layers of multi-head attention and convolution subcomponents. Each transformed token is then passed through a \emph{head} and \emph{tail} MLP to produce two position-specific representations. A bi-affine operation is performed between each \emph{head} and \emph{tail} representation with respect to each relation's embedding matrix, producing a pair-wise relation affinity tensor. Finally, the scores for cells corresponding to the same entity pair are pooled with a separate LogSumExp operation for each relation to get a final score. The colored tokens illustrate calculating the score for a given pair of entities; the model is only given entity information when pooling over mentions. \label{fig:transformer_all_pairs}}
  \end{minipage}
\end{figure}

\subsection {Inputs}
Our model takes in a sequence of $N$ token embeddings in $\mathbb{R}^d$. Because the Transformer has no innate notion of token position, the model relies on positional embeddings which are added to the input token embeddings.\footnote{Though our final model incorporates some convolutions, we retain the position embeddings.} We learn the position embedding matrix $P^{m\times d}$ which contains a separate $d$ dimensional embedding for each position, limited to $m$ possible positions. Our final input representation for token $x_i$ is:
\begin{align*}
x_i=s_i+p_i
\end{align*}
where $s_i$ is the token embedding for $x_i$ and $p_i$ is the positional embedding for the $i$th position. If $i$ exceeds $m$, we use a randomly initialized vector in place of $p_i$.

We tokenize the text using byte pair encoding (BPE) \citep{gage1994new,sennrich2015neural}. The BPE algorithm constructs a vocabulary of sub-word pieces, beginning with single characters. Then, the algorithm iteratively merges the most frequent co-occurring tokens into a new token, which is added to the vocabulary. This procedure continues until a pre-defined vocabulary size is met. 

BPE is well suited for biological data for the following reasons. First, biological entities often have unique mentions made up of meaningful subcomponents, such as \emph{1,2-dimethylhydrazine}. Additionally, tokenization of chemical entities is challenging, lacking a universally agreed upon algorithm \citep{krallinger2015chemdner}. As we demonstrate in \S \ref{sec:ctd-results}, the sub-word representations produced by BPE allow the model to formulate better predictions, likely due to better modeling of rare and unknown words.

\subsection {Transformer}\label{sec:model_multihead_attention}
We base our token encoder on the Transformer self-attention model \citep{vaswani2017attention}. The Transformer is made up of $B$ blocks. Each Transformer block, which we denote $\mathrm{Transformer}_k$, has its own set of parameters and is made up of two subcomponents: multi-head attention and a series of convolutions\footnote{The original Transformer uses feed-forward connections, i.e. width-1 convolutions, whereas we use convolutions with width $>$ 1.}. The output for token $i$ of block $k$, $b_i^{(k)}$, is connected to its input $b_i^{(k-1)}$ with a residual connection \citep{he2016deep}. Starting with $b_i^{(0)} = x_i$:
\begin{align*}
b_i^{(k)}=b_i^{(k-1)}+\mathrm{Transformer}_k(b_i^{(k-1)})
\end{align*}

\subsubsection{Multi-head Attention\label{sec:multihead-attention}}
Multi-head attention applies self-attention multiple times over the same inputs using separately normalized parameters (attention heads) and combines the results, as an alternative to applying one pass of attention with more parameters. The intuition behind this modeling decision is that dividing the attention into multiple heads make it easier for the model to learn to attend to different types of relevant information with each head. The self-attention updates input $b_i^{(k-1)}$ by performing a weighted sum over all tokens in the sequence, weighted by their importance for modeling token $i$. 

Each input is projected to a key $k$, value $v$, and query $q$, using separate affine transformations with ReLU activations \citep{glorot2011deep}. Here, $k$, $v$, and $q$ are each in $\mathbb{R}^{\frac{d}{H}}$ where $H$ is the number of heads. The attention weights $a_{ijh}$ for head $h$ between tokens $i$ and $j$ are computed using scaled dot-product attention:
\begin{align*}
a_{ijh} &= \sigma\left(\frac{q_{ih}^T k_{jh}}{\sqrt{d}}\right) \\ 
o_{ih} &= \sum_j v_{jh} \odot a_{ijh} \\
\end{align*}
with $\odot$ denoting element-wise multiplication and $\sigma$ indicating a softmax along the $j$th dimension. The scaled attention is meant to aid optimization by flattening the softmax and better distributing the gradients \citep{vaswani2017attention}.

The outputs of the individual attention heads are concatenated, denoted $[\cdot;\cdot]$, into $o_i$. All layers in the network use residual connections between the output of the multi-headed attention and its input. Layer normalization \citep{ba2016layer}, denoted LN$(\cdot)$, is then applied to the output. 
 \begin{align*}
o_i &= [o_1;...;o_h] \\
m_i &= \mathrm{LN}(b_i^{(k-1)} + o_i)
\end{align*}

\subsubsection{Convolutions\label{sec:model_feedforward}}

The second part of our Transformer block is a stack of convolutional layers. The sub-network used in \citet{vaswani2017attention} uses two width-1 convolutions. We add a third middle layer with kernel width 5, which we found to perform better. Many relations are expressed concisely by the immediate local context, e.g. \emph{Michele's husband Barack}, or \emph{labetalol-induced hypotension}. Adding this explicit n-gram modeling is meant to ease the burden on the model to learn to attend to local features. We use $C_w(\cdot)$ to denote a convolutional operator with kernel width $w$. Then the convolutional portion of the transformer block is given by:
\begin{align*}
t_i^{(0)} &= \mathrm{ReLU}(C_1(m_i)) \\
t_i^{(1)} &= \mathrm{ReLU}(C_5(t_i^{(0)})) \\
t_i^{(2)} &= C_1(t_i^{(1)})
\end{align*}
Where the dimensions of $t_i^{(0)}$ and $t_i^{(1)}$ are in $\mathbb{R}^{4d}$ and that of $t_i^{(2)}$ is in $\mathbb{R}^{d}$.

\subsection {Bi-affine Pairwise Scores}

We project each contextually encoded token $b_i^{(B)}$ through two separate MLPs to generate two new versions of each token corresponding to whether it will serve as the first (head) or second (tail) argument of a relation:
\begin{align*}
e^{head}_i &= W_{head}^{(1)}(\mathrm{ReLU}(W_{head}^{(0)} b_i^{(B)})) \\
e^{tail}_i &= W_{tail}^{(1)}(\mathrm{ReLU}(W_{tail}^{(0)} b_i^{(B)}))
\end{align*}
We use a bi-affine operator to calculate an $N \times L \times N$ tensor $A$ of pairwise affinity scores, scoring each (head, relation, tail) triple:
\begin{align*}
A_{ilj} = (e^{head}_i L)e^{tail}_j
\end{align*}
where $L$ is a $d \times L \times d$ tensor, a learned embedding matrix for each of the $L$ relations. In subsequent sections we will assume we have transposed the dimensions of $A$ as $d \times d \times L$ for ease of indexing.

\subsection {Entity Level Prediction \label{sec:entity_pred}}
Our data is weakly labeled in that there are labels at the entity level but not the mention level, making the problem a form of strong-distant supervision \citep{distant_supervision}. In distant supervision, edges in a knowledge graph are heuristically applied to sentences in an auxiliary unstructured text corpus --- often applying the edge label to all sentences containing the subject and object of the relation. Because this process is imprecise and introduces noise into the training data, methods like multi-instance learning were introduced \citep{riedel2010modeling, surdeanu2012multi}. In multi-instance learning, rather than looking at each distantly labeled mention pair in isolation, the model is trained over the aggregate of these mentions and a single update is made. More recently, the weighting function of the instances has been expressed as neural network attention \citep{verga-mccallum:2016:W16-13, lin2016neural, yaghoobzadeh-adel-schutze:2017:EACLlong}. 

We aggregate over all representations for each mention pair in order to produce per-relation scores for each entity pair. For each entity pair $(p^{head}, p^{tail})$, let $P^{head}$ denote the set of indices of mentions of the entity $p^{head}$, and let $P^{tail}$ denote the indices of mentions of the entity $p^{tail}$. Then we use the LogSumExp function to aggregate the relation scores from $A$ across all pairs of mentions of $p^{head}$ and $p^{tail}$:
\begin{align*}
scores(p^{head}, p^{tail}) = \log\sum_{\mathclap{\substack{i \in P^{head}\\ j \in P^{tail}}}}\exp(A_{ij})
\end{align*}
The LogSumExp scoring function is a smooth approximation to the max function and has the benefits of aggregating information from multiple predictions and propagating dense gradients as opposed to the sparse gradient updates of the max \citep{das-EtAl:2017:EACLlong1}.

\subsection{Named Entity Recognition}
\label{sec:model_NER}
In addition to pairwise relation predictions, we use the Transformer output $b_i^{(B)}$ to make entity type predictions. We feed $b_i^{(B)}$ as input to a linear classifier which predicts the entity label for each token with per-class scores $c_i$:
\begin{align*}
c_i = W^{(3)} b_i^{(B)}
\end{align*}
We augment the entity type labels with the BIO encoding to denote entity spans. We apply tags to the byte-pair tokenization by treating each sub-word within a mention span as an additional token with a corresponding B- or I- label.


\subsection{Training}
We train both the NER and relation extraction components of our network to perform multi-class classification using maximum likelihood, where NER classes $y_i$ or relation classes $r_i$ are conditionally independent given deep features produced by our model with probabilities given by the softmax function. In the case of NER, features are given by the per-token output of the transformer:
\begin{align*}
\frac{1}{N}\sum_{i=1}^N \log P(y_i\mid b_i^{(B)} )
\end{align*}
In the case of relation extraction, the features for each entity pair are given by the LogSumExp over pairwise scores described in \S~\ref{sec:entity_pred}. For $E$ entity pairs, the relation $r_i$ is given by:
\begin{align*}
\frac{1}{E}\sum_{i=1}^E \log P(r_i\mid scores(p^{head}, p^{tail}))
\end{align*}
We train the NER and relation objectives jointly, sharing all embeddings and Transformer parameters. To trade off the two objectives, we penalize the named entity updates with a hyperparameter $\lambda$.

\section{Results}
We evaluate our model on three datasets: The Biocreative V Chemical Disease Relation benchmark (CDR), which models relations between chemicals and diseases (\S \ref{sec:cdr_data}); the Biocreative VI ChemProt benchmark (CPR), which models relations between chemicals and proteins (\S \ref{sec:chemprot_results}); and a new, large and accurate dataset we describe in \S \ref{sec:ctd_results} based on the human curation in the Chemical Toxicology Database (CTD), which models relationships between chemicals, proteins and genes. 

The CDR dataset is annotated at the level of paper abstracts, requiring consideration of long-range, cross sentence relationships, thus evaluation on this dataset demonstrates that our model is capable of such reasoning. We also evaluate our model's performance in the more traditional setting which does not require cross-sentence modeling by performing experiments on the CPR dataset, for which all annotations are between two entity mentions in a single sentence. Finally, we present a new dataset constructed using strong-distant supervision (\S \ref{sec:entity_pred}), with annotations at the document level. This dataset is significantly larger than the others, contains more relation types, and requires reasoning across sentences.
\subsection{Chemical Disease Relations Dataset\label{sec:cdr_data}}

The Biocreative V chemical disease relation extraction (CDR) dataset\footnote{\url{http://www.biocreative.org/}} \citep{li2016biocreative,wei2016assessing} was derived from the Comparative Toxicogenomics Database (CTD), which curates interactions between genes, chemicals, and diseases \citep{davis2008comparative}. CTD annotations are only at the document level and do not contain mention annotations. The CDR dataset is a subset of these original annotations, supplemented with human annotated, entity linked mention annotations. The relation annotations in this dataset are also at the document level only. 

\subsubsection{Data Preprocessing \label{sec:data_processing}}
The CDR dataset is concerned with extracting only chemically-induced disease relationships (drug-related side effects and adverse reactions) concerning the most specific entity in the document. For example \emph{tobacco causes cancer} could be marked as false if the document contained the more specific \emph{lung cancer}. This can cause true relations to be labeled as false, harming evaluation performance. To address this we follow \cite{gu2016chemical, gu2017chemical} and filter hypernyms according to the hierarchy in the MESH controlled vocabulary\footnote{\url{https://www.nlm.nih.gov/mesh/download/2017MeshTree.txt}}. All entity pairs within the same abstract that do not have an annotated relation are assigned the NULL label.

In addition to the gold CDR data, \citet{peng2016improving} add 15,448 PubMed abstracts annotated in the CTD dataset. We consider this same set of abstracts as additional training data (which we subsequently denote +Data). Since this data does not contain entity annotations, we take the annotations from Pubtator \citep{pubtator2013}, a state of the art biological named entity tagger and entity linker. See \S \ref{sec:cdr_impl_details} for additional data processing details. In our experiments we only evaluate our relation extraction performance and all models (including baselines) use gold entity annotations for predictions. 

The byte pair vocabulary is generated over the training dataset --- we use a budget of 2500 tokens when training on the gold CDR data, and a larger budget of 10,000 tokens when including extra data described above Additional implementation details are included in Appendix \ref{appendix}.

\begin{table}[htp]
\begin{tabular}{llll}
    Data split & Docs & Pos & Neg \\ \hline \hline
    Train &  500 & 1,038 & 4,280 \\
    Development & 500 &  1,012 & 4,136 \\
    Test &  500 & 1,066 & 4,270 \\
    CTD  &  15,448 & 26,657 & 146,057 \\
  \end{tabular}
  \caption{Data statistics for the CDR Dataset and additional data from CTD. Shows the total number of abstracts, positive examples, and negative examples for each of the data set splits.  \label{data:stats}}
\end{table}


\subsubsection{Baselines}
We compare against the previous best reported results on this dataset not using knowledge base features.\footnote{The highest reported score is from \cite{peng2016improving}, but they use explicit lookups into the CTD knowledge base for the existence of the test entity pair.} Each of the baselines are ensemble methods for within- and cross-sentence relations that make use of additional linguistic features (syntactic parse and part-of-speech). \citet{gu2017chemical} encode mention pairs using a CNN while \citet{zhou2016exploiting} use an LSTM. Both make cross-sentence predictions with featurized classifiers.

\subsubsection{Results}
In Table \ref{results:CDR} we show results outperforming the baselines despite using no linguistic features. We show performance averaged over 20 runs with 20 random seeds as well as an ensemble of their averaged predictions. We see a further boost in performance by adding weakly labeled data. Table \ref{results:cdr_ablation} shows the effects of ablating pieces of our model. `CNN only' removes the multi-head attention component from the transformer block, `no width-5' replaces the width-5 convolution of the feed-forward component of the transformer with a width-1 convolution and `no NER' removes the named entity recognition multi-task objective (\S \ref{sec:model_NER}). 

\begin{table}
\centering
\begin{tabular}{llll}
    Model & P & R & F1 \\ \hline \hline
    \citet{gu2016chemical}      &  62.0  & 55.1   & 58.3 \\
    \citet{zhou2016exploiting}  &  55.6  & 68.4   & 61.3  \\
    \citet{gu2017chemical}      &  55.7  & 68.1   & 61.3 \\
        \hline
    BRAN                        &  55.6 & 70.8 &  \textbf{62.1} $\pm$ 0.8  \\
    + Data                      &  64.0  & 69.2 & \textbf{66.2} $\pm$ 0.8 \\
    \hline
    BRAN(ensemble)              &  63.3 & 67.1 &  {65.1}  \\
    + Data                      &  65.4 & 71.8 &  \textbf{68.4}  \\
    \hline 
  \end{tabular}
  \caption{Precision, recall, and F1 results on the Biocreative V CDR Dataset.\label{results:CDR}}
\end{table}

\begin{table}
\centering
\begin{tabular}{llll}
    Model & P & R & F1 \\ \hline \hline
    BRAN (Full)                 &  55.6 & 70.8 &  \textbf {62.1} $\pm$ 0.8  \\
    -- CNN only                 &  43.9  & 65.5 & 52.4 $\pm$ 1.3 \\
    -- no width-5               &  48.2  & 67.2 & 55.7 $\pm$ 0.9  \\
    -- no NER                   &  49.9  & 63.8 & 55.5 $\pm$ 1.8 \\
        \hline 
  \end{tabular}
  \caption{Results on the Biocreative V CDR Dataset showing precision, recall, and F1 for various model ablations. \label{results:cdr_ablation}}
\end{table}

\subsection{Chemical Protein Relations Dataset \label{sec:chemprot_results}}
To assess our model's performance in settings where cross-sentence relationships are not explicitly evaluated, we perform experiments on the Biocreative VI ChemProt dataset (CDR) \cite{biocreative6}. This dataset is concerned with classifying into six relation types between chemicals and proteins, with nearly all annotated relationships occurring within the same sentence.

\subsubsection{Baselines}
We compare our models against those competing in the official Biocreative VI competition \citep{liu_bc6}. We compare to the top performing team whose model is directly comparable with ours --- i.e. used a single (non-ensemble) model trained only on the training data (many teams use the development set as additional training data). The baseline models are standard state of the art relation extraction models: CNNs and Gated RNNs with attention. Each of these baselines uses mention-specific features encoding relative position of each token to the two target entities being classified, whereas our model aggregates over all mention pairs in each sentence. It is also worth noting that these models use a large vocabulary of pre-trained word embeddings, giving their models the advantage of far more model parameters, as well as additional information from unsupervised pre-training.

\subsubsection{Results}
In Table \ref{results:bc6} we see that even though our model forms all predictions simultaneously between all pairs of entities within the sentence, we are able to outperform state of the art models classifying each mention pair independently. The scores shown are averaged across 10 runs with 10 random seeds. Interestingly, our model appears to have higher recall and lower precision, while the baseline models are both precision-biased, with lower recall. This suggests that combining these styles of model could lead to further gains on this task.

\label{sec:bc6_data}
\begin{table}
\centering
\begin{tabular}{llll}
    Model & P & R & F1 \\ \hline \hline
    CNN$\dagger$                         & 50.7  & 43.0 & 46.5 \\
    GRU+Attention$\dagger$               & 53.0  & 46.3 & 49.5 \\
    BRAN                                 & 48.0  & 54.1 & \textbf{50.8} $\pm$ .01  \\
    \hline 
  \end{tabular}
  \caption{Precision, recall, and F1 results on the Biocreative VI Chem-Prot Dataset. $\dagger$ denotes results from \citet{liu_bc6}\label{results:bc6}}
\end{table}
\hfill
\subsection{New CTD Dataset\label{sec:ctd_results}}

\subsubsection{Data}
Existing biological relation extraction datasets including both CDR (\S \ref{sec:cdr_data}) and CPR (\S \ref{sec:chemprot_results}) are relatively small, typically consisting of hundreds or a few thousand annotated examples. Distant supervision datasets apply document-independent, entity-level annotations to all sentences leading to a large proportion of incorrect labels. Evaluations on this data involve either very small (a few hundred) gold annotated examples or cross validation to predict the noisy, distantly applied labels \cite{mallory2015large,quirk-poon:2017:EACLlong, peng2017cross}. 

We address these issues by constructing a new dataset using strong-distant supervision containing document-level annotations. The Comparative Toxicogenomics Database (CTD) curates interactions between genes, chemicals, and diseases. Each relation in the CTD is associated with a disambiguated entity pair and a PubMed article where the relation was observed. 

To construct this dataset, we collect the abstracts for each of the PubMed articles with at least one curated relation in the CTD database. As in \S \ref{sec:cdr_data}, we use PubTator to automatically tag and disambiguate the entities in each of these abstracts. If both entities in the relation are found in the abstract, we take the (abstract, relation) pair as a positive example. The evidence for the curated relation could occur anywhere in the full text article, not just the abstract. Abstracts with no recovered relations are discarded. All other entity pairs with valid types and without an annotated relation that occur in the remaining abstracts are considered negative examples and assigned the NULL label. We additionally remove abstracts containing greater than 500 tokens\footnote{We include scripts to generate the unfiltered set of data as well to encourage future research}. This limit removed about 10\% of the total data including numerous extremely long abstracts. The average token length of the remaining data was \~230 tokens. With this procedure, we are able to collect 166,474 positive examples over 13 relation types, with more detailed statistics of the dataset listed in Table \ref{data:ctd_stats}. 

We consider relations between chemical-disease, chemical-gene, and gene-disease entity pairs downloaded from CTD\footnote{\url{http://ctdbase.org/downloads/}}. We remove inferred relations (those without an associated PubMed ID) and consider only human curated relationships. Some chemical-gene entity pairs were associated with multiple relation types in the same document. We consider each of these relation types as a separate positive example. 

The chemical-gene relation data contains over 100 types organized in a shallow hierarchy. Many of these types are extremely infrequent, so we map all relations to the highest parent in the hierarchy, resulting in 13 relation types. Most of these chemical-gene relations have an increase and decrease version such as increase\_expression and decrease\_expression. In some cases, there is also an affects relation (affects\_expression) which is used when the directionality is unknown. If the affects version is more common, we map decrease and increase to affects. If affects is less common, we drop the affects examples and keep the increase and decrease examples as distinct relations, resulting in the final set of 10 chemical-gene relation types.

\begin{table}[t]
\begin{tabular}{l|lll}
    Types & Docs & Pos & Neg \\ \hline \hline
    Total &  68,400 & 166,474 & 1,198,493 \\
    Chemical/Disease &  64,139 & 93,940 & 571,932 \\
    Chemical/Gene  & 34,883 & 63,463 & 360,100 \\
    Gene/Disease &  32,286 & 9,071 & 266,461 \\
  \end{tabular}
  \caption{Data statistics for the new CTD dataset. \label{data:ctd_stats}}
\end{table}

\begin{table}[t]
\hspace*{-.4cm}
  \centering
\begin{tabular}{l|lll}
                 & Train &  Dev & Test \\ \hline \hline
    \textbf {Total} & 120k  & 15k &  15k \\
    \hline
    \textbf {Chemical / Disease } \\
    \hline
    marker/mechanism           &  41,562 & 5,126 & 5,167  \\
    therapeutic                &  24,151 & 2,929 & 3,059 \\
    \hline 
    \textbf {Gene / Disease } \\
    \hline
    marker/mechanism           &  5,930  & 825  & 819 \\
    therapeutic                &  560   & 77   & 75 \\
    \hline 
    \textbf {Chemical / Gene } \\
    \hline
    increase\_expression            &  15,851 & 1,958 & 2,137 \\
    increase\_metabolic\_proc &  5,986  & 740  & 638   \\
    decrease\_expression            &  5,870  & 698  & 783  \\
    increase\_activity              &  4,154  & 467  & 497   \\
    affects\_response               &  3,834   & 475  & 508  \\
    decrease\_activity              &  3,124  & 396  & 434  \\
    affects\_transport              &  3,009  & 333  & 361   \\
    increase\_reaction              &  2,881  & 367  & 353  \\
    decrease\_reaction              &  2,221  & 247  & 269  \\
    decrease\_metabolic\_proc &  798   & 100  & 120  \\
    \hline
  \end{tabular}
  \caption{Data statistics for the new CTD dataset broken down by relation type. The first column lists relation types separated by the types of the entities. Columns 2--4 show the number of positive examples of that relation type. 
  \label{data:ctf_rel_stats}}
\end{table}

\subsubsection{Results\label{sec:ctd-results}}
In Table \ref{ctf-results} we list precision, recall and F1 achieved by our model on the CTD dataset, both overall and by relation type. Our model predicts each of the relation types effectively, with higher performance on relations with more support. 

In Table \ref{results:ctd_ner} we see that our sub-word BPE model out-performs the model using the Genia tokenizer \cite{geniatagger} even though our vocabulary size is one-fifth as large. We see a 1.7 F1 point boost in predicting Pubtator NER labels for BPE. This could be explained by the increased out-of-vocabulary (OOV) rate for named entities. Word training data has 3.01 percent OOV rate for tokens with an entity. The byte pair-encoded data has an OOV rate of 2.48 percent. Note that in both the word-tokenized and byte pair-tokenized data, we replace tokens that occur less than five times with a learned UNK token.

\begin{table}[t]
  \centering
\begin{tabular}{l|lll}
                 &  P & R & F1\\ \hline \hline
    \textbf {Total} \\
    \hline
    Micro F1              & 44.8 & 50.2 & 47.3 \\
    Macro F1              & 34.0 & 29.8 & 31.7 \\
    \hline
    \textbf {Chemical / Disease } \\
    \hline
    marker/mechanism            & 46.2 & 57.9 & 51.3 \\
    therapeutic                 & 55.7 & 67.1 & 60.8 \\
    \hline 
    \textbf {Gene / Disease } \\
    \hline
    marker/mechanism           & 42.2 & 44.4 & 43.0 \\
    therapeutic                & 52.6 & 10.1 & 15.8 \\
    \hline 
    \textbf {Chemical / Gene } \\
    \hline
    increases\_expression        & 39.7 & 48.0 & 43.3 \\
    increases\_metabolic\_proc    & 26.3 & 35.5 & 29.9 \\
    decreases\_expression        & 34.4 & 32.9 & 33.4 \\
    increases\_activity          & 24.5 & 24.7 & 24.4 \\
    affects\_response           & 40.9 & 35.5 & 37.4 \\
    decreases\_activity          & 30.8 & 19.4 & 23.5 \\
    affects\_transport           & 28.7 & 23.8 & 25.8 \\
    increases\_reaction          & 12.8 & 5.6 & 7.4 \\
    decreases\_reaction          & 12.3 & 5.7 & 7.4 \\
    decreases\_metabolic\_proc    & 28.9 & 7.0 & 11.0 \\
    \hline
  \end{tabular}
  \caption{BRAN precision, recall and F1 results for the full CTD dataset by relation type. The model is optimized for micro F1 score across all types.
  \label{ctf-results}}
\end{table}

\begin{figure}[t]
\hspace*{-.3cm}
    \includegraphics[scale=.46]{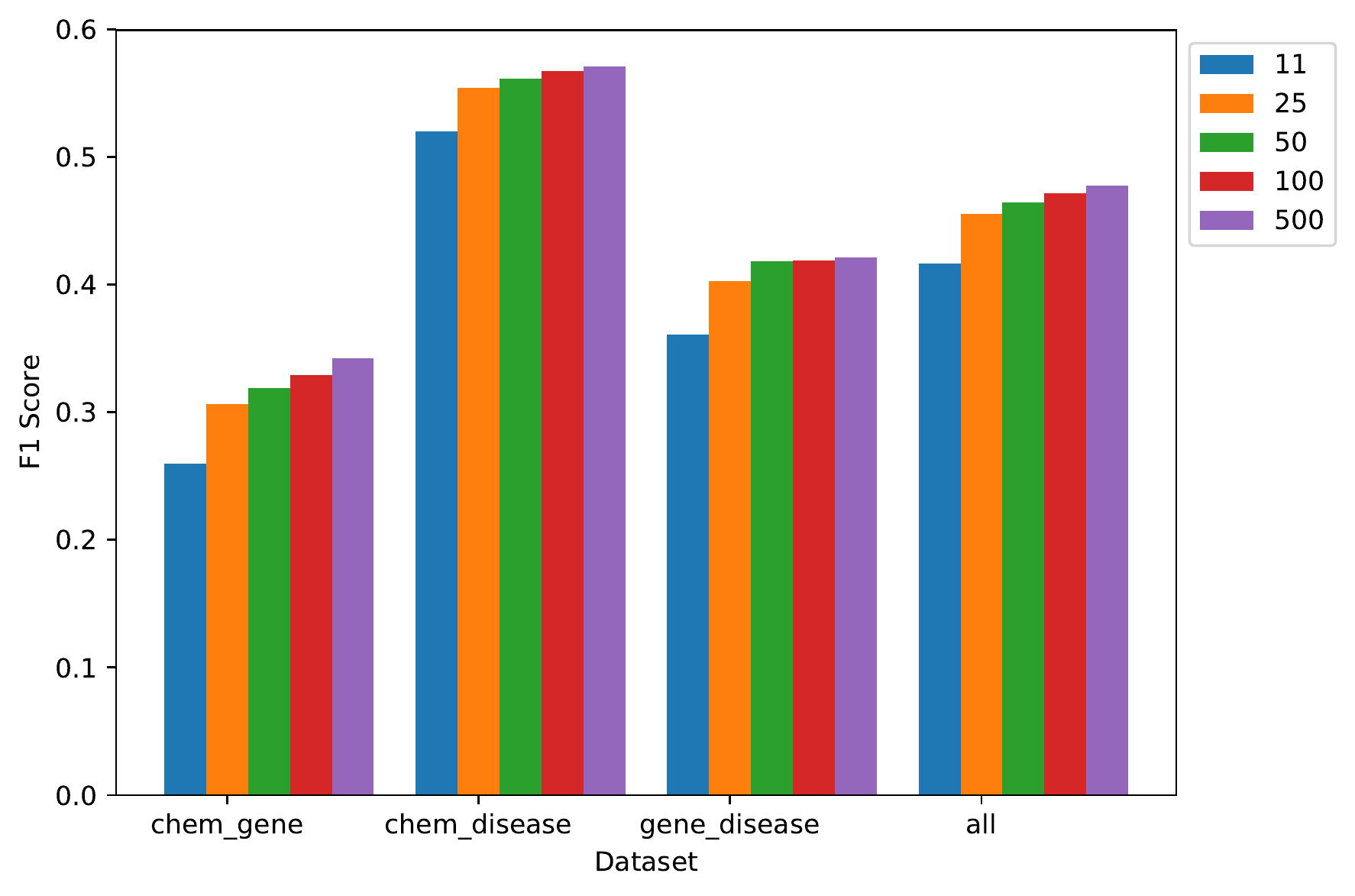}
    \caption{Performance on the CTD dataset when restricting candidate entity pairs by distance. The x-axis shows the coarse-grained relation type. The y-axis shows F1 score. Different colors denote maximum distance cutoffs. \label{fig:ctd_bar}}
\end{figure}

Figure \ref{fig:ctd_bar} depicts the model's performance on relation extraction as a function of distance between entities. For example, the blue bar depicts performance when removing all entity pair candidates (positive and negative) whose closest mentions are more than 11 tokens apart. We consider removing entity pair candidates with distances of 11, 25, 50, 100 and 500 (the maximum document length). The average sentence length is 22 tokens. We see that the model is not simply relying on short range relationships, but is leveraging information about distant entity pairs, with accuracy increasing as the maximum distance considered increases. Note that all results are taken from the same model trained on the full unfiltered training set.

\begin{table}
\centering
\begin{tabular}{llll}
    \bf Model & \bf P & \bf R & \bf F1 \\ \hline \hline
    \multicolumn{4}{l}{{\bf Relation extraction}} \\
    \hline
    Words   &  44.9 & 48.8 & 46.7 $\pm$ 0.39  \\
    BPE  &  44.8 & 50.2 & \textbf{47.3} $\pm$ 0.19  \\
    \hline
    \multicolumn{4}{l}{{\bf NER}} \\
    \hline
    Words   &  91.0  & 90.7 &  90.9  $\pm$ 0.13  \\
    BPE  &  91.5 &  93.6 & \textbf{92.6} $\pm$ 0.12  \\
    \hline 
  \end{tabular}
  \caption{Precision, recall, and F1 results for CTD named entity recognition and relation extraction, comparing BPE to word-level tokenization.\label{results:ctd_ner}}
\end{table}
\hfill

\section{Related work}
Relation extraction is a heavily studied area in the NLP community. Most work focuses on news and web data \citep{ACE2004,riedel2010modeling,hendrickx2009semeval}.\footnote{And TAC KBP: \url{https://tac.nist.gov}} Recent neural network approaches to relation extraction have focused on CNNs \citep{dossantos-xiang-zhou:2015:ACL-IJCNLP,zeng-EtAl:2015:EMNLP} or LSTMs \citep{miwa-bansal:2016:P16-1, verga-EtAl:2016:N16-1,zhou-EtAl:2016:P16-2} and replacing stage-wise information extraction pipelines with a single end-to-end model \citep{miwa-bansal:2016:P16-1, Ammar2017ai2semeval,li2017neural}. These models all consider mention pairs separately. 

There is also a considerable body of work specifically geared towards supervised biological relation extraction including protein-protein \citep{pyysalo2007bioinfer,poon2014distant, mallory2015large}, drug-drug \citep{segurabedmar-martinez-herrerozazo:2013:SemEval-2013}, and chemical-disease \citep{gurulingappa2012development,li2016biocreative} interactions, and more complex events \citep{kim2008corpus, riedel-EtAl:2011:BioNLP-ST}. Our work focuses on modeling relations between chemicals, diseases, genes and proteins, where available annotation is often at the document- or abstract-level, rather than the sentence level. 

Some previous work exists on cross-sentence relation extraction. \citet{swampillai-stevenson:2011:RANLP} and \citet{quirk-poon:2017:EACLlong} consider featurized classifiers over cross-sentence syntactic parses. Most similar to our work is that of \citet{peng2017cross}, which uses a variant of an LSTM to encode document-level syntactic parse trees. Our work differs in three key ways. First, we operate over raw tokens negating the need for part-of-speech or syntactic parse features which can lead to cascading errors. We also use a feed-forward neural architecture which encodes long sequences far more efficiently compared to the graph LSTM network of \citet{peng2017cross}. Finally, our model considers all mention pairs simultaneously rather than a single mention pair at a time.

We employ a bi-affine function to form pairwise predictions between mentions. Such models have also been used for knowledge graph link prediction \citep{rescal,li-EtAl:2016:P16-14}, with variations such as restricting the bilinear relation matrix to be diagonal \citep{distmult} or diagonal and complex \citep{trouillon2016complex}. Our model is similar to recent approaches to graph-based dependency parsing, where bilinear parameters are used to score head-dependent compatibility \citep{kipperwasser-goldberg-parser, dozat2016deep}.

\section{Conclusion}
We present a bi-affine relation attention network that simultaneously scores all mention pairs within a document. Our model performs well on three datasets, including two standard benchmark biological relation extraction datasets and a new, large and high-quality dataset introduced in this work. Our model out-performs the previous state of the art on the Biocreative V CDR dataset despite using no additional linguistic resources or mention pair-specific features. 

Our current model predicts only into a fixed schema of relations given by the data. However, this could be ameliorated by integrating our model into open relation extraction architectures such as Universal Schema \cite{riedel2013relation, verga2016multilingual}. Our model also lends itself to other pairwise scoring tasks such as hypernym prediction, co-reference resolution, and entity resolution. We will investigate these directions in future work. 

\section*{Acknowledgments}
We thank Ofer Shai and the Chan Zuckerberg Initiative / Meta data science team for helpful discussions. We also thank Timothy Dozat and Kyubyong Park for releasing their code. 
This material is based upon work supported in part by the Center for Data Science and the Center for Intelligent Information Retrieval, in part by the Chan Zuckerberg Initiative under the project “Scientific Knowledge Base Construction, in part by the National Science Foundation under Grant No. DMR-1534431 and  IIS-1514053. ES is supported by an IBM PhD Fellowship Award. Any opinions, findings and conclusions or recommendations expressed in this material are those of the authors and do not necessarily reflect those of the sponsor.

\bibliography{sources}
\bibliographystyle{acl_natbib}

\clearpage
\newpage

\appendix

\section{Implementation Details \label{appendix}}

The model is implemented in Tensorflow \citep{tensorflow2015-whitepaper} and trained on a single TitanX gpu. The number of transformer block repeats is $B=2$ . We optimize the model using Adam \cite{kingma2014adam} with best parameters chosen for $\epsilon$, $\beta_1$, $\beta_2$ chosen from the development set. The learning rate is set to $0.0005$ and batch size 32. In all of our experiments we set the number of attention heads to $h=4$. 

We clip the gradients to norm 10 and apply noise to the gradients \cite{neelakantan2015adding}. We tune the decision threshold for each relation type separately and perform early stopping on the development set. We apply dropout \cite{srivastava2014dropout} to the input layer randomly replacing words with a special UNK token with keep probability $.85$. We additionally apply dropout to the input $T$ (word embedding + position embedding), interior layers, and final state. At each step, we randomly sample a positive or negative (NULL class) minibatch with probability $0.5$.

\subsection{Chemical Disease Relations Dataset\label{sec:cdr_impl_details}}
Token embeddings are pre-trained using skipgram \cite{mikolov2013efficient} over a random subset of 10\% of all PubMed abstracts with window size 10 and 20 negative samples. We merge the train and development sets and randomly take 850 abstracts for training and 150 for early stopping. Our reported results are averaged over 10 runs and using different splits. All baselines train on both the train and development set. Models took between 4 and 8 hours to train. 

$\epsilon$ was set to 1e-4, $\beta_1$ to .1, and $\beta_2$ to 0.9. Gradient noise $\eta=.1$. Dropout was applied to the word embeddings with keep probability 0.85, internal layers with 0.95 and final bilinear projection with 0.35 for the standard CRD dataset experiments. When adding the additional weakly labeled data: word embeddings with keep probability 0.95, internal layers with 0.95 and final bilinear projection with 0.5. 

\subsection{Chemical Protein Relations Dataset\label{sec:bc6_impl_details}}
We construct our byte-pair encoding vocabulary using a budget of 7500. The dataset contains annotations for a larger set of relation types than are used in evaluation. We train on only the relation types in the evaluation set and set the remaining types to the Null relation. The embedding dimension is set to 200 and all embeddings are randomly initialized. $\epsilon$ was set to 1e-8, $\beta_1$ to .1, and $\beta_2$ to 0.9. Gradient noise $\eta=1.0$. Dropout was applied to the word embeddings with keep probability 0.5, internal layers with 1.0 and final bilinear projection with 0.85 for the standard CRD dataset experiments.

\subsection{Full CTD Dataset\label{sec:ctd_impl_details}}
We tune separate decision boundaries for each relation type on the development set. For each prediction, the relation type with the maximum probability is assigned. If the probability is below the relation specific threshold, the prediction is set to NULL. We use embedding dimension 128 with all embeddings randomly initialized. Our byte pair encoding vocabulary is constructed with a budget of 50,000. Models took 1 to 2 days to train.

$\epsilon$ was set to 1e-4, $\beta_1$ to .1, and $\beta_2$ to 0.9. Gradient noise $\eta=.1$.Dropout was applied to the word embeddings with keep probability 0.95, internal layers with 0.95 and final bilinear projection with 0.5

\end{document}